# Surprise! You've Got Some Explaining to Do…

**Meadhbh I. Foster (meadhbh.foster@ucdconnect.ie)**
**Mark T. Keane (mark.keane@ucd.ie)**
Department of Computer Science & Informatics, University College Dublin
Belfield, Dublin 4, Ireland

**Abstract**

Why are some events more surprising than others? We propose that events that are more difficult to explain are those that are more surprising. The two experiments reported here test the impact of different event outcomes (Outcome-Type) and task demands (Task) on ratings of surprise for simple story scenarios. For the Outcome-Type variable, participants saw outcomes that were either *known* or *less-known* surprising outcomes for each scenario. For the Task variable, participants either answered comprehension questions or provided an explanation of the outcome. Outcome-Type reliably affected surprise judgments; known outcomes were rated as less surprising than less-known outcomes. Task also reliably affected surprise judgments; when people provided an explanation it lowered surprise judgments relative to simply answering comprehension questions. Both experiments thus provide evidence on this less-explored explanation aspect of surprise, specifically showing that ease of explanation is a key factor in determining the level of surprise experienced.

**Keywords:** Surprise; explanation; comprehension, coherence

## Introduction

Life is full of surprises, from bumping into a friend from home while on holidays, to arriving at a surprise party, to opening an amazing birthday gift, or hitting paydirt on that 100-1 racehorse. Surprise has been researched since Darwin's time, perhaps because it involves an interesting mixture of emotion and cognition. Originally, it was conceived of as a "basic emotion" (see Darwin, 1872; Ekman & Friesan, 1971; Izard, 1977; Plutchik, 1991; Tomkins, 1962), though more recently it has been re-appraised as a cognitive state (Kahneman & Miller, 1986; Maguire, Maguire & Keane, 2011) because, unlike most emotions, it can either be positively or negatively valenced (Ortony & Turner, 1990). Although surprise clearly involves an emotional reaction (often accompanied by a startle response) it may also serve a strategic, cognitive goal, as it directs attention to explaining why the surprising event occurred and to learning for the future (see e.g., Maguire et al., 2011; Ranganath & Rainer, 2003). Accordingly, in Artificial Intelligence (AI), surprise is seen as a candidate mechanism for identifying learning events in agent architectures (Bae & Young, 2008, 2009; Macedo & Cardoso, 2001; Macedo, Reisenzein & Cardoso, 2004).

Imagine that you walk into your house and the walls have changed color from the color they were this morning. If you have no explanation for this turn of events then you would probably be surprised by this outcome[1]. Many outcomes are surprising, the question is why? Our answer is that outcomes are surprising when they are hard to explain. Specifically, that surprise *is a meta-cognitive sense of the amount of explanatory, mental work that was carried out to establish coherence between unfolding events in the world.*

To illustrate the point, consider different scenarios for the "re-decoration surprising outcome". If I had left a team of decorators in my house that morning, I would clearly be less surprised by my walls being re-painted, because I had planned for that to occur. If no decorators were contracted, then I would be really surprised at this outcome, because no obvious explanation is forthcoming. However, if my wife and friends have been smirking at me for weeks (the way they do when they throw surprise parties) I would be less surprised because I can explain it as a prank. The experience of surprise will gradually increase across these scenarios as they move from being thoroughly-explainable (contracted decorators) to potentially explainable (smirking friends) to thoroughly-unexplainable (no decorators or smirking) because people have to carry out more explanatory, mental work to establish the coherence of these unfolding events.

In theories of surprise, one group of theorists have focussed on the properties of surprising outcomes, characterising them as low-probability events, disconfirmed expectations or schema-discrepant events (e.g., Meyer, Reisenzein & Schützwohl, 1997; Reisenzein & Studtmann, 2007; Schützwohl & Reisenzein, 1999). Another group of theorists have stressed the importance of (often retrospective) sense-making and the integration of the surprising outcome to make it cohere with previous events (Kahneman & Miller, 1986; Maguire & Keane, 2006; Maguire et al., 2011). Theoretically, we are more aligned with the latter than with the former group; the main novelty in our approach being its emphasis on the meta-cognitive, explanatory aspects of the sense-making process. Adopting this meta-cognitive, explanatory approach suggests that experienced surprise may differ (a) for different classes of surprising outcomes (i.e., known versus less-known outcomes) and (b) under different task demands (i.e., being explicitly asked to explain a surprising outcome or not).

---

[1] We use the term "surprising outcome" in this paper to denote the target surprising event because traditional terminology is too theory-laden; for instance, "unexpected event" suggests one had expectations about the event when this is not always the case, and "abnormal event" presupposes some unspecified normative standard.

## Classes of Surprising Outcomes

Viewing surprise from an explanation-perspective, suggests that outcomes may vary in their surprisingness because some are more well-known (directly or vicariously) than others. Intuitively, losing your wallet and losing your belt (that you put on your jeans this morning) are outcomes that could both surprise you during your day. We could call "losing your wallet" a *known surprising outcome* as it is an experience that people often discuss with one another, suggesting that most people have several "ready-made" explanations for it (see also Schank, 1986); that I left it in a shop, that I dropped it or that I was pickpocketed. In contrast, "losing your belt" is a *less-known surprising outcome*, suggesting perhaps that there are few or no "ready-made" explanations for it[2]. We predict that differences in the explanation spaces for these different classes of outcomes will result in different amounts of mental work to make them coherent and, thus, result in different levels of experienced surprise. Traditional probabilistic accounts would recast this known/less-known dimension as some variation of subjective probability, making parallel predictions about levels of surprise. However, obviously, we do not think that subjective probability is the key predictor of behaviour; indeed, in related work where it has been explicitly assessed, it has been shown not to accurately predict levels of surprise (see Maguire et al., 2011, Experiment 1).

## Explanation Task

Viewing surprise as a meta-cognitive effect suggests that if we ask people to explicitly explain the surprising outcome, they will be less surprised than if they receive task demands that are less directed toward explanation (e.g., comprehension questions about the scenario). If people are in "explanation mode" then clearly they should expend less mental effort in explaining the surprising event and hence, other things being equal, should experience less surprise relative to being in some "non-explanation mode". Should such explanation-effects occur, they can probably be explained in some ad hoc fashion by probabilistic accounts; however, we cannot see how a probabilistic account would lead one to perform such a test.

## Experiment 1

To test these predictions, we asked people to make surprise ratings about the outcomes of simple story scenarios describing everyday events. Some outcomes were known surprising outcomes, others were less-known surprising outcomes (see operational definitions in *Materials*). The task demands were varied by asking participants to either produce the answer to two short comprehension questions about that story or to produce an explanation for why that outcome may have occurred. So, the experiment involved a 2 x 2 design with Task (explanation vs. comprehension) as a between-subjects variable and Outcome-Type (known vs. less-known) as a within-subjects variable. The questions asked for the comprehension task were very simple, using information clearly and unambiguously presented in the text given to participants (e.g., "*Where is [character's name]?*").

First, it was predicted that scenarios involving *known surprising outcomes* would be rated as less surprising than those with the *less-known surprising outcomes*; as explanations (or partial explanations) for the former would be available for use in making the outcome cohere with the rest of the scenario. Second, it was also predicted that the task demand to find an explanation would result in lower surprise ratings for outcomes, relative to the task demand of answering comprehension questions on the same stories. We made no specific predictions about whether these two variables would interact.

## Method

**Participants and Design** Forty UCD students (12 male, 28 female) with a mean age of 21.2 years ($SD$ = 2.07, range = 19-29) took part voluntarily in this study. Informed consent was obtained prior to the experiment. Participants were randomly assigned to one of two conditions in a 2 (between-subjects; Task: explanation versus comprehension) x 2 (within-subjects; Outcome-Type: known versus less-known) mixed-measures design.

**Materials** A material set was created consisting of simple story scenarios with outcomes that were designed to involve known or less-known surprising outcomes (see Table 1). The type of outcome was operationally defined using (a) a pre-test sorting task by an independent group of raters and (b) Latent Semantic Analysis (LSA) scores of coherence.

*For the sorting task definition*, 20 story scenarios were presented in a pre-test to independent raters ($N$ = 10). The raters were assigned to two groups: one group received half the scenarios with a known surprising outcome and the other half of the scenarios with a less-known surprising outcome, and the second group received the opposite. Each rater saw only one outcome for a given scenario. They were asked to determine if a given scenario has an outcome that "falls within the range of reasonable outcomes to the scenario" (i.e., known surprising outcome) or whether it "falls less within the range of reasonable outcomes to the scenario" (i.e., less-known surprising outcome). Of the 20 stories, the raters consistently deemed 9 stories to have separable known and less-known surprising outcomes (Fleiss' kappa showed substantial agreement, κ = .68, Landis & Koch, 1977).

*For the coherence-score definition*, the known and less-known variants of these 9 stories were scored using LSA. In discourse research (cf., Graesser & McNamara, 2011), the explanatory coherence of texts is often operationalized by using latent semantic analysis (LSA) scores, where higher LSA scores indicate that the one text is more coherent than another (Landauer & Dumas, 1996, 1997). For the selected

---

[2] The only plausible explanation we could garner was leaving your belt at the security area in an airport.

9 stories used in the experiment, the scenarios with the known outcomes were scored higher ($M = .62$, $SD = .2$) than their matched counterparts with less-known outcomes ($M = .53$, $SD = .21$), a difference that was statistically reliable, $F(1,8) = 9.47$, $p = .015$, $\eta_p^2 = .54$.

Four material sets were created. Each of these comprised all nine scenarios, with either four scenarios with known surprising outcomes and five with less-known surprising outcomes, or five scenarios with known surprising outcomes and four with less-known surprising outcomes. As expected, the four material sets used proved to have no effect on subsequent surprise judgments, so these results are not reported in the following analyses ($p > .12$)

The order of presentation of these stories was randomised for each participant. Stories were presented on separate pages of a booklet, which began with the appropriate task instructions (explanation or comprehension). Each story was presented on a separate page with the scenario setting on the top of the page, followed by the outcome (known/less-known), the statement of the task (comprehension or explanation) and a 7-point scale on which to rate the suprisingness of the outcome (1: not surprised to 7: very surprised).

**Procedure and Scoring** Participants were asked to read nine stories and to judge the surprisingness of their outcomes (see Table 1). For the Task variable, the participants in the explanation condition were asked to produce the first explanation they could think of for why the outcome may have occurred, before rating it for surprise; in the comprehension condition the participants were asked to answer two simple comprehension questions about the scenario, before rating it for surprise. For each story, the first question was about the story setting, and the second question was about the outcome.

Table 1: Sample scenario used in Experiment 1.

| Setting | Rebecca is on the beach. She goes for a swim in the water. | |
|---|---|---|
| **Outcome** | **Known** After she dries herself off she notices that her skin has turned red. | **Less-known** After she dries herself off she notices that her skin has turned turquoise. |

Prior to the experiment, we conducted a pre-test ($N = 4$) to verify that there was no significant difference in the average time taken to produce an explanation compared to that taken to answer the two short comprehension questions; time taken to do one task or the other were not reliably different ($t(2) = -1.41$, $p = .29$, explanation $M = 6.5$ minutes; comprehension $M = 7.5$ minutes). Two measures were recorded: (a) the 7-point scale rating of surprise, and (b) the explanations produced by participants for each scenario in the explanation group. Finally, prior to data analysis one participant (2.5% of the data) was discarded because they failed to follow the instructions given.

**Results and Discussion**

Overall, the results confirmed the predictions that Outcome-Type and Task both impact people's perceptions of surprise. The intuition that known outcomes are less surprising than less-known outcomes was confirmed, as was the prediction that instructions to explain the outcome would reduce the overall perception of surprise. So, for example, though both outcomes were deemed to be surprising, the lost-wallet type of scenario was found to be less surprising than the lost-belt type of scenario. No reliable interaction was found between the two variables.

**Surprise Judgments** A two-way ANOVA confirmed that participants judged stories with known outcomes ($M = 3.92$, $SD = 1.18$) to be less surprising than those with less-known outcomes ($M = 5.73$, $SD = 0.95$), $F(1,37) = 128.82$, $p < .001$, $\eta_p^2 = .78$, see Figure 1. We maintain that this Outcome-Type effect occurs because known outcomes have associated "ready-made" explanations that are recruited quickly and easily to explain the outcome, lowering surprise ratings. In contrast, stories with less-known outcomes have few "ready-made" explanations to be recruited, so the outcome is harder to explain, resulting in relatively higher surprise ratings.

There was also a significant main effect of Task, $F(1,37) = 10.18$, $p = .003$, $\eta_p^2 = .22$, indicating that the explanation group judged the outcomes to be less surprising ($M = 4.40$, $SD = 1.03$) than the comprehension group ($M = 5.27$, $SD = 0.62$). This effect occurs because in 'explanation mode' participants find explanations more easily and, hence, for meta-cognitive reasons, their perception of surprise decreases. No interaction between the two variables was found, $F(1,37) = 0.00$, $p = .98$, $\eta_p^2 < .001$.

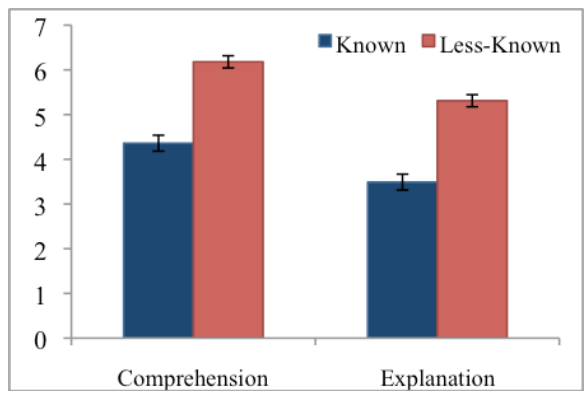

Figure 1: Mean surprise ratings for both levels of Outcome-Type (known vs. less-known) and Task (explanation vs. comprehension) in Experiment 1

**Explanations** The explanations provided by the participants in the explanation group provide a key piece of converging evidence for the view that known outcomes differ from less-known outcomes. Participants' explanations for each scenario were recorded and classified to identify the most common or dominant explanation for a given scenario. We then carried out a by-materials analysis of the scenarios using the frequency of this dominant explanation as the dependent measure. The ANOVA revealed a main effect of Outcome-Type, in which dominant explanations were found to be more frequently produced to known outcomes ($M = 5.44$, $SD = 1.59$) than less-known outcomes ($M = 4$, $SD = 1.32$), $F(1,8) = 6.76$, $p = .03$, $\eta_p^2 = .46$. So, participants agree more about the explanations for known outcomes than they do for less-known outcomes, showing that the explanation spaces for these classes of outcomes differ.

## Experiment 2

Our second experiment attempted to replicate the effects found for Outcome-Type and Task, while adding a manipulation to the setting (Setting-Type) designed to elicit counterfactuals, to test another potential aspect of surprise.

Kahneman & Tversky (1982; Kahneman & Miller, 1986) proposed that "abnormal events" (our "surprising outcomes") will seem more abnormal if contrasting counterfactual alternatives are highly available; that is, the abnormal event (i.e., losing your wallet) will appear more abnormal if the contrasting counterfactual (i.e., the normal event of "having your wallet") is highly available. Kahneman & Miller also propose that the availability of the normal event (the counterfactual) can provide an explanation for the abnormal event (the factual one), as people often use the difference between the two events to find an explanation. So, in theory, the elicitation of such counterfactuals could reduce the perceived surprise of an outcome, as it could provide a "quick and dirty" explanation of the surprising outcome. However, this prediction assumes that the counterfactual-inspired explanation is always used (which may not be a given). The literature on counterfactuals (Byrne, 2002; Kahneman & Tversky, 1982) shows that they tend to be elicited when scenarios mention non-routine events (e.g., if you are told Jack had a car crash when he did not take his usual route home, people naturally draw on the counterfactual scenario of Jack taking his usual route home to find an explanation), though this is not always the case (e.g., Dixon & Byrne, 2011). So, in this experiment, in addition to the original settings used in Experiment 1 (none), to elicit counterfactuals we changed the setting in the scenarios to stress that the event was either routine (usual) or non-routine (exceptional; see Table 2) for the actor involved.

So, the final design for this experiment manipulated Task (comprehension versus explanation), Outcome-Type (known versus less-known) and Setting-Type (none, usual or exceptional).

## Method

**Participants and design** Sixty UCD students (27 male, 33 female) with a mean age of 20.95 years ($SD = 4.228$, range = 18-44) took part voluntarily in this study. Informed consent was obtained prior to the experiment. Participants were randomly assigned to one of two conditions in a 2 (between-subjects; Task: comprehension versus explanation) x 2 (within-subjects; Outcome-type: known versus less-known) x 3 (within-subjects; Setting-Type: none, usual, exceptional) mixed-measures design.

**Procedure and Scoring** As in Experiment 1, participants were asked to read nine stories and to judge the surprisingness of their outcomes. Rather than asking participants how surprised they would be "if this event occurred" (as they were in Experiment 1), they were asked to judge how surprised they would be by the event "if they were the character described". For the Setting-Type variable, the events in the story setting (a) gave no hint as to whether they were routine or not (none), (b) were said to be regular or routine (usual), or (c) said to be non-usual or non-routine (exceptional). For the Outcome-Type variable, the participants saw either a known or less-known surprising outcome for each story; only one outcome and one setting was seen by each participant for each story (see Table 2 for an example of the materials used). The LSA scores for the three variants of the setting, none, usual and exceptional showed no main effect of this Setting-Type variable ($p > .59$).

Table 2: Sample scenario used in Experiment 2

|  | **None** | **Usual** | **Exceptional** |
|---|---|---|---|
| **Sentence 1** | Lorna is in an ethnic restaurant. | Lorna is in her favourite ethnic restaurant that she has often gone to before. | Lorna is in a new ethnic restaurant that she has never gone to before. |
| **Sentence 2** | She has ordered her food and, after a while, the waiter brings it to her. | | |
| **Outcome** | **Known**: When she asks for a knife she is told that they have none. | | **Less-known**: When she asks for a knife she is brought a banana. |

Six material sets were created. Each of these comprised all 9 scenarios, with three variants of each setting type (none, usual, exceptional). Of these, either four scenarios were presented with known surprising outcomes and five with less-known surprising outcomes, or five scenarios with known surprising outcomes and four with less-known surprising outcomes. As expected, the six material sets had no effect on subsequent surprise judgments, so were not included as a variable in the reported analyses ($p > .5$).

The order of presentation of these stories was randomised anew for each participant. Stories were presented sentence by sentence on a desktop computer-screen as participants pressed the spacebar, with each sentence appearing below the preceding one on the screen, until the outcome was presented. At this point, the participants in the explanation condition were instructed to "*type in the first explanation you can think of for why this outcome may have occurred:*" and the participants in the comprehension condition saw and answered sequentially two simple comprehension questions about the story. One of these questions was about the information provided in the setting, and the other was about information provided in the outcome. Neither of these questions drew the participants' attention to the Setting-Type variable, *per se*. Initially, the participants in this condition saw the first question and, after providing an answer, they pressed the return key, this first question disappeared and the second question appeared. After the explanation/comprehension step, all participants pressed the return key and the question "*If you were [character's name], how surprised would you be by this outcome?*" On presentation of this question, participants indicated on a 7-point scale their surprise judgment (1: not surprised, to 7: very surprised). Three measures were recorded: (a) the 7-point rating of surprise, (b) the response time from the time of seeing the outcome sentence to the time in which the surprise judgment was made[3], and (c) the explanations produced by each participant for each scenario. Finally, prior to data analysis, four participants (6.7% of the data) was discarded for failing to follow the instructions given.

**Results and Discussion**

Overall, the results confirmed the predictions that known surprising outcomes and the adoption of an "explanation-mode" decreased the perception of surprise; however, there was no strong evidence for a counterfactual effect.

**Surprise Judgments** A three-way ANOVA confirmed that participants judged known outcomes to be less surprising ($M = 4.51$, $SD = 1.11$) than less-known outcomes ($M = 6.21$, $SD = .75$), showing a main effect of Outcome-Type, $F(1,54) = 92.46$, $p < .001$, $\eta_p^2 = .63$. There was also a significant main effect of Task, $F(1,54) = 4.64$, $p = .036$, $\eta_p^2 = .08$. indicating that participants judged the outcomes of scenarios to be more surprising when they had answering comprehension questions, ($M = 5.56$, $SD = .63$) as opposed to providing explanations for them ($M = 5.09$, $SD = .85$; see Figure 2). However, there was no main effect of Setting-Type, $F(2,108) = .002$, $p = .998$, $\eta_p^2 < .001$, no interaction between Outcome-Type and Setting-Type, $F(2,108) = 2.78$, $p = .07$, $\eta_p^2 = .05$, and no reliable 2-way interactions between the variables (all $F$s < 1).

**Explanations** Again the frequency with which the most dominant explanation was chosen by the explanation group was calculated for each scenario. A two-way, by-materials ANOVA showed a main effect of Outcome-Type, in which participants were more likely to produce the same dominant explanation for a known surprising outcome ($M = 7.89$, $SD = 3.26$) than for a less-known outcome ($M = 5.22$, $SD = 2.63$), $F(1,8) = 6.09$, $p = .039$, $\eta_p^2 = .43$. So, again, participants seem to have a greater degree of shared knowledge in the explanation of known outcomes than they do for less-known outcomes, showing that the explanation spaces for these classes of outcomes differ.

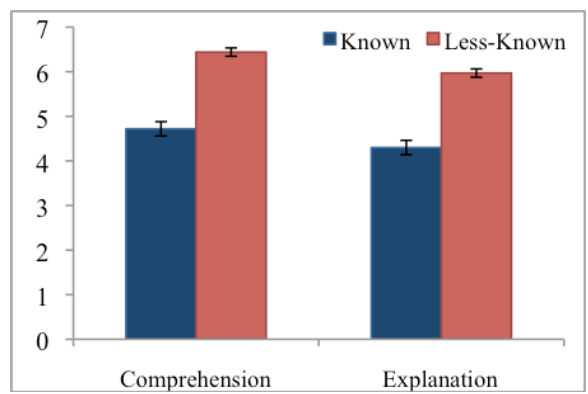

Figure 2: Mean surprise ratings for both levels of Outcome-Type (known vs. less-known) and Task (explanation vs. comprehension) in Experiment 2

**General Discussion**

Overall, the experiments showed that known surprising outcomes are perceived as less surprising than less-known outcomes for the same scenarios, presumably because they are easier to explain. The task of explaining itself was also found to significantly reduce surprise ratings relative to answering comprehension questions in both experiments, again demonstrating how explanation may be the key factor in determining the level of surprise experienced. Finally, the explanations produced by participants were found to be more homogeneous for known outcomes than for less-known outcomes; that is, there seems to be a shared dominant explanation used to explain known outcomes, that is less present in the case of less-known outcomes. We believe that these results provide converging evidence for an explanation-based account of surprise. Indeed, taken together, the combined effects on surprise found here strongly suggests that surprise may be a metacognitive effect (see Müller & Stahlberg, 2007; Sanna & Lundberg, 2012; Touroutoglou & Efklides, 2010), with perceived surprise reflecting the ease or difficulty of explaining the surprising event.

However, little evidence was found for the counterfactual effect tested for in Experiment 2 (see the Setting-type variable). Both Kahneman & Miller's Norm Theory (1986) and Teigen & Keren's Contrast Hypothesis (2003) seem to predict that the ready availability of counterfactuals may influence the degree of surprise experienced; norm theory proposes that counterfactuals are used to explain why the event occurred, while the contrast hypothesis proposes that what was expected to occur (the events these counterfactuals elicit) is contrasted with the outcome to determine the level of surprise. There are several possible reasons for this prediction failure; it could be that our manipulation was not notable enough to elicit counterfactuals (though prior research would suggest otherwise), or it could be that counterfactuals were generated but not used for explanation, or not considered as good-enough explanations. Of course, it could also be the case that the prediction is just wrong.

---
[3] Unreported in this paper for space reasons.

The current work also has implications for AI approaches to agent architectures, where it has been proposed that surprise might be used to identify learning events (e.g, Macedo & Cardoso, 2001; Macedo, Reisenzein & Cardoso, 2004). This proposal looks like it could be useful, once it is tempered by some consideration of the degree of surprise entailed and the ease of producing an explanation. The current work suggests that both of these aspects of the surprise process can differ considerably and, as such, would deliver very different learning outcomes for an agent.